\pdfoutput=1

\documentclass[11pt]{article}

\usepackage[preprint]{coling}

\usepackage{hyperref}
\usepackage{times}
\usepackage{latexsym}
\usepackage{tabularx} 
\usepackage{tablefootnote}
\usepackage{longtable}
\usepackage{multirow}                 
\usepackage{multicol}                 
\usepackage{multirow}                
\usepackage{makecell}                 
\usepackage{booktabs}
\usepackage{amssymb}
\usepackage[T1]{fontenc}

\usepackage[utf8]{inputenc}

\usepackage{microtype}

\usepackage{inconsolata}

\usepackage{graphicx}

\title{Teaching-Inspired Integrated Prompting Framework: A Novel Approach for Enhancing Reasoning in Large Language Models}

\author{
  Wenting Tan\thanks{Work done during internship at Youdao, NetEase.} \\
  The University of Hong Kong \\
  \texttt{wt212796@connect.hku.hk}\\\And
  Dongxiao Chen \\
  Youdao, NetEase, Inc. \\
  \texttt{chendx@rd.netease.com} \\\And
  Jieting Xue \\
  Youdao, NetEase, Inc. \\
  \texttt{xuejt@rd.netease.com} \AND
  Zihao Wang \\
  Youdao, NetEase, Inc. \\
  \texttt{wangzh03@rd.netease.com} \\\And
  Taijie Chen \\
  The University of Hong Kong \\
  \texttt{ctj21@connect.hku.hk} \\
}

\begin{document}
\maketitle
\begin{abstract}
Large Language Models (LLMs) exhibit impressive performance across various domains but still struggle with arithmetic reasoning tasks. Recent work shows the effectiveness of prompt design methods in enhancing reasoning capabilities. However, these approaches overlook crucial requirements for prior knowledge of specific concepts, theorems, and tricks to tackle most arithmetic reasoning problems successfully. To address this issue, we propose a novel and effective Teaching-Inspired Integrated Framework, which emulates the instructional process of a teacher guiding students. This method equips LLMs with essential concepts, relevant theorems, and similar problems with analogous solution approaches, facilitating the enhancement of reasoning abilities. Additionally, we introduce two new Chinese datasets, MathMC and MathToF, both with detailed explanations and answers. Experiments are conducted on nine benchmarks which demonstrates that our approach improves the reasoning accuracy of LLMs. With GPT-4 and our framework, we achieve new state-of-the-art performance on four math benchmarks (AddSub, SVAMP, Math23K and AQuA) with accuracies of 98.2\% (+3.3\%), 93.9\% (+0.2\%), 94.3\% (+7.2\%) and 81.1\% (+1.2\%). Our data and code are available at \href{https://github.com/SallyTan13/Teaching-Inspired-Prompting}{https://github.com/SallyTan13/Teaching-Inspired-Prompting}.

\end{abstract}

\section{Introduction}

Large Language Models (LLMs) have made significant strides in the field of Natural Language Processing (NLP), demonstrating outstanding performance across various tasks \cite{devlin2018bert,brown2020language,chowdhery2022palm}. Nonetheless, handling reasoning tasks effectively remains a challenge for LLMs. Evidence suggests that simply scaling up the model size does not provide an adequate solution to this issue \cite{rae2021scaling,srivastava2022beyond}. 

To address this issue, a series of new prompting methods are proposed to enhance reasoning in LLMs. The effectiveness of these methods is substantiated by extensive experiments. Chain-of-Thought (CoT) prompting \cite{wei2022chain}, which mimics the human approach to solving multi-step problems by providing LLMs with few-shot exemplars including intermediate reasoning steps. Based on CoT, subsequent studies have further refined this method and improved the performance, such as Zero-shot-CoT \cite{kojima2022large}, Complexity-based CoT \cite{fu2022complexity} and Least-to-Most Prompting \cite{zhou2022least}. Self-consistency (SC) is also a breakthrough method that replaces the greedy decoding strategy used in CoT but samples various reasoning paths and selects the answer with the highest consistency \cite{wang2022self}. From another perspective, MathPrompter \cite{imani2023mathprompter} and Program of Thoughts (PoT) prompting \cite{chen2022program} empower LLMs to generate programming language statements, enabling them to provide more accurate solutions for complex mathematical calculations.

While prompting methods mentioned above greatly improve the reasoning performance of LLMs, they miss the crucial need for a strong grasp of concepts, theorems, and strategies. Firstly, the knowledge repository of LLMs may be incomplete, lacking enough conceptual and theoretical foundation to tackle certain arithmetic reasoning problems. Secondly, lack of familiarity with specific problem-solving strategies may lead LLMs to incorrect preconditions, resulting in inaccurate final answers despite correct intermediate calculations. These challenges also arise in the process of human practice and problem-solving. Like teachers who provide foundational concepts and examples for students before practice, LLMs require educational-sourced information to ensure accurate reasoning and solutions. Therefore, drawing inspiration from traditional teaching methods, we propose a Teaching-Inspired Integrated Prompting Framework. This framework imitates the guidance provided by teachers to students by drawing from curated educational databases to offer clear, problem-specific concepts or theorems as background knowledge. It also presents similar problems with easy-to-learn solution ideas as examples. Additionally, it incorporates answer double-checking and selection mechanisms to enhance the overall reasoning ability of LLMs.

Existing arithmetic benchmarks contain a limited number of Multiple-Choice and True-or-False questions. Hence, we create two Chinese mathematical datasets called MathMC and MathToF comprising 1,000 Multiple-Choice math problems and 1,000 True-or-False math problems respectively with answers and detailed rationales. 

Our approach is evaluated on nine benchmarks, including six English datasets, one Chinese dataset, and two datasets we created. These experiments are conducted on GPT-3.5-Turbo \cite{ouyang2022training} and GPT-4 \cite{openaigpt} respectively. Experimental results demonstrate that the reasoning performances of both language models on nine benchmarks are improved.

Our main contributions are as follows:
\begin{itemize}
    \item A novel teaching-inspired integrated prompting framework is proposed to improve the reasoning capabilities of LLMs.
    \item Two Chinese arithmetic datasets (MathMC and MathToF) with answers and detailed rationales are constructed for further facilitating the study of arithmetic reasoning tasks.\footnote{We will make the dataset publicly when the paper is published.} 
    \item Comprehensive experiments show the effectiveness of our proposed integrated framework, and it achieves new state-of-the-art performance on four benchmarks.
\end{itemize}

\section{Related Work}
\subsection{In-context Learning}
In-context learning (ICL) has emerged as a successful and widely adopted approach for NLP tasks. It enables language models to learn and make predictions based on a few examples \cite{dong2022survey}. Unlike supervised learning, ICL does not rely on vast amounts of data and resources for training and fine-tuning language models which makes LLMs easier to apply to various tasks as a service \cite{sun2022black}. By designing the template or format of demonstration and select more relevant exemplars \cite{wei2022chain, fu2022complexity, chen2022program, xiong2023dq}, the effectiveness of utilizing LLMs to address complex reasoning tasks, such as arithmetic reasoning and commonsense reasoning, has significantly improved. 

\subsection{Reasoning with Prompting}
\noindent\textbf{Chain-of-Thought Based Prompting.} As for CoT prompting \cite{wei2022chain}, the language model is given a few exemplars with intermediate reasoning steps so that it can offer intermediate steps when solving multi-step problems. Building upon this, Kojima et al. \shortcite{kojima2022large} introduced Zero-shot-CoT, a method that simplifies the human annotation process by replacing few-shot examples with the prompt "Let's think step by step". Subsequently, Complexity-based CoT was introduced \cite{fu2022complexity}, targeting example selection for multi-step reasoning and demonstrating that inputs with more reasoning chains yield superior performance. This concept is expanded to output selection, favoring results with more reasoning steps. Active Prompting \citep{diao2023active} leverages uncertainty metrics, aiding the selection of the most informative and important questions for annotation. Furthermore, Self-Consistency \cite{wang2022self} focuses on refining the original greedy decoding strategy in CoT by sampling different reasoning paths and selecting the most frequently occurring answer.

\noindent\textbf{Program Based Prompting.} Unlike CoT prompting, Chen et al. \shortcite{chen2022program} proposed Program-of-Chain. This approach generates Python programs using LLMs and employs a Python interpreter to compute results, addressing LLMs' limitations in complex calculations and error tendencies. Building on this, Imani et al. \shortcite{imani2023mathprompter} introduce MathPrompter, which also leverages LLMs to generate Python programs and algebraic expressions.
\begin{table*}[!t]
\setlength\tabcolsep{3pt}
    \centering
    \begin{tabular}{l|ccccccc}
        \Xhline{1pt}
         & Arithmetic & Algebra & Geometry & Statistics & Reasoning & Others & Total\\    
         \hline
         MathMC & 619 & 113 & 227 & 27 & 7 & 7 & 1000\\
         \hline
         MathToF & 675 & 61 & 197 & 37 & 13 &17 & 1000\\
        \Xhline{1pt}
    \end{tabular} 
    \caption{Question Types of MathMC and MathToF}
    \label{tbl:tip_our_datasets}
\end{table*}

\noindent\textbf{External Knowledge Enhanced Reasoning.}
Despite the impressive knowledge base and generative capabilities of Large Language Models (LLMs), they still often generate hallucinations or erroneous information. Recent studies demonstrate that augmenting prompts with external knowledge for in-context learning, can enhance their reasoning abilities \cite{rubin2022learning}. Lu et al. \shortcite{lu2022dynamic} proposed a dynamic prompt learning method by policy gradients learning to select in-context examples from training data. Similarly, a post-processing method, Rethinking with Retrieval \citep{he2022rethinking},  retrieves external knowledge corresponding to a set of reasoning steps of CoT, giving more faithful explanations. To enhance the retrieval of relevant external information, planning and self-enhancement techniques are employed \citep{Lee2024PlanRAGAP,Asai2023SelfRAGLT}.

\section{MathMC and MathToF}
We create two datasets, MathMC and MathToF, featuring 1,000 Chinese mathematical Multiple-Choice and 1,000 Chinese True-or-False questions, accompanied by detailed explanations addressing the lack of diverse question types in existing Chinese arithmetic datasets.

In constructing these datasets, we initiated by gathering a set of 4,000 elementary school-level seed Multiple-Choice questions and 4,000 seed True-or-False questions spanning grades 1 to 6, focusing on math problems typically encountered in grades 4 to 6. These questions are then carefully filtered and proofread to ensure a broad coverage of knowledge points in each question. Through this rigorous filtering and selection process, we create a final dataset of 1,000 Multiple-Choice questions and 1,000 True-or-False questions, each meticulously labeled with answers and explanations. Two datasets feature a wide range of question types, including arithmetic, algebra, geometry, statistics, reasoning, and others. Specific question types statistics are shown in Table \ref{tbl:tip_our_datasets}. 

Appendix \ref{data-appendix} presents details of our created two datasets and two sample Multiple-Choice and True-or-False questions from our datasets.

\section{Teaching-Inspired Integrated Prompting Framework}

\begin{figure*}[!t]
\centering
\includegraphics[trim=0 12cm 0 3cm,width=0.9\textwidth]{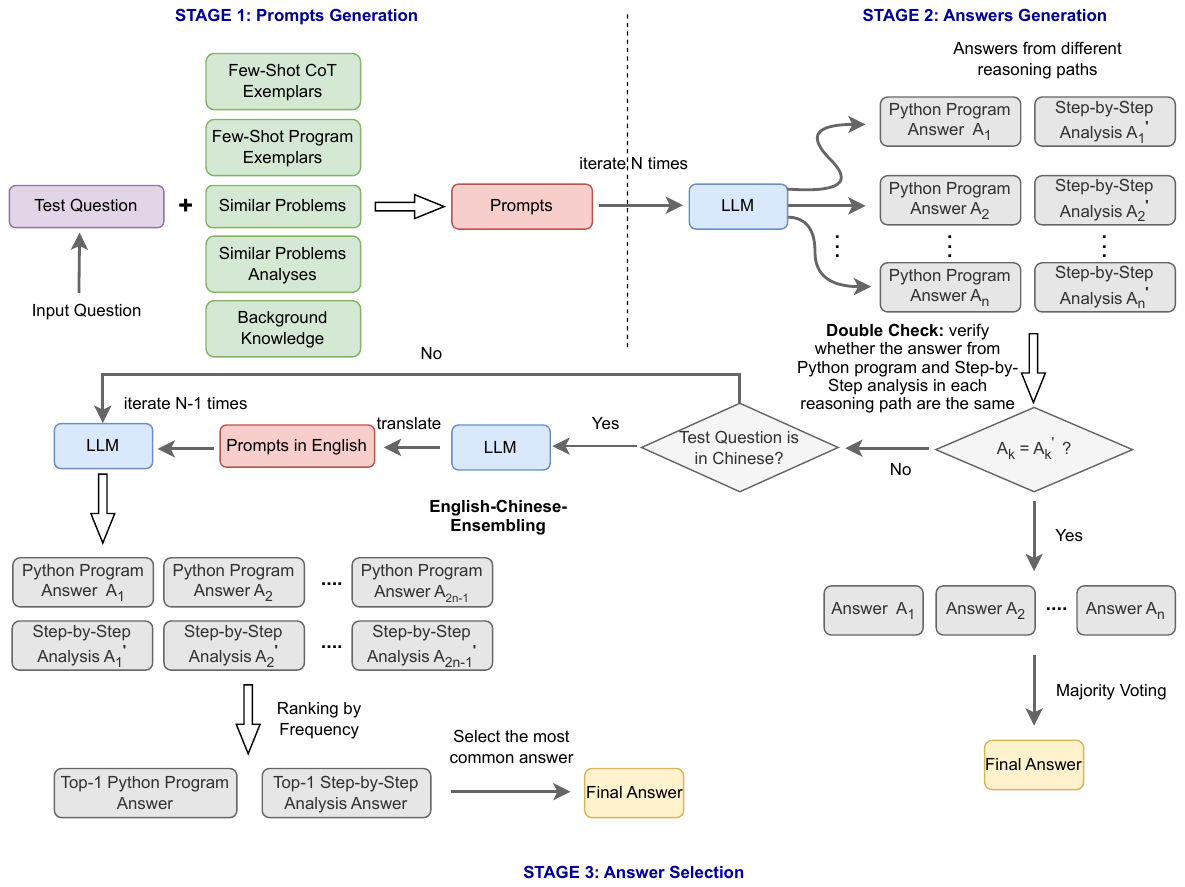} 
\caption{Architecture of our Teaching-Inspired Integrated Prompting Framework}
\label{fig3}
\end{figure*}

We construct a prompting framework, as illustrated in Figure \ref{fig3} which consists of three stages.
\subsection{Teaching-Inspired Prompts Generation}
Prompts are generated by drawing inspiration from traditional pedagogical methods, emphasizing the use of educational sources. Students begin with foundational theories and concepts from textbooks and workbooks to deeply understand problem principles, then apply these through extensive exercises and examples. To enhance the capability of LLMs to solve mathematical reasoning problems, we adapt the aforementioned teaching strategy to reasoning tasks, feeding the LLMs with similar problems and the essential background knowledge (e.g. theorems, concepts, and term definitions) required to solve the specific problem. 

Figure \ref{fig_sim_bg}(a) illustrates the process of obtaining similar problems. We tokenize the test problem, preserving special math expressions, and use BM25 \cite{robertson2009probabilistic} to retrieve a set of candidate problems, $\mathbb{P}$. Identical problems and those differing only in numerical values are excluded. Candidates are ranked by their Longest Common Subsequence (LCS) length with the test problem.

Figure \ref{fig_sim_bg}(b) describes background knowledge acquisition. We tokenize the test problem and analyses, constructing a token set $\mathbb{T}$ by removing stopwords and operands. An LLM aids in extracting key knowledge points and uncertain theorems. These tokens serve as queries to retrieve relevant theorems and conceptual knowledge from a knowledge database, yielding background knowledge candidates, $\mathbb{K}$. Candidates are ranked by LCS length, with the top three selected. Similar to obtaining similar problems, the LCS length between each candidate $k_i$ and the combined text is computed. Top-3 candidates are selected based on LCS length.

 Therefore, prompts are mainly composed of three elements: few-shot CoT + PoT exemplars (2 cases) (one case = question + CoT exemplars + Python program exemplars + answer), similar questions and their analyses, and background knowledge. These prompts assist LLMs in generating the intermediate steps of the final answer and devising the Python program needed to solve the problem.
\subsection{Answers Generation}
We employ the self-consistency method, allowing LLMs to iterate N times, generating N different paths (problem-solving strategies) for the answers.
\subsection{Answers Selection}
\textbf{Double-Check Verification.} We initially compare the results generated by each pathway in the N possible solution paths, i.e., verifying if the outputs of the Python programs align with the corresponding step-by-step answers. This comparison process double-checks the computation results, thereby enhancing the trustworthiness of the final answer. If all paths yield consistent answers, the most frequent answer from the consistent answers is chosen as the output via majority voting. Otherwise, the LLM is tasked to provide N-1 with more answers to the problem. After that, the process transitions to the Further Selection stage. The inclusion of Python programs and the implementation of a double-check-verification strategy reduce the probability of simple calculation errors and enhance the reliability of the language model.

\noindent\textbf{English-Chinese Ensembling.} In the additional N-1 solution requests, if the problem is in Chinese, we instruct the language model to translate the test problem, the background knowledge, and similar problems into English before generating solutions. This approach is adopted since LLM might not fully understand certain Chinese expressions, and translation can aid in generating accurate results.

\noindent\textbf{Further Selection.} We evaluate the frequency of the most common answer in both the Python program outputs (code-ans) and the results derived from step-by-step solutions (step-by-step-ans). If the most frequent answer from the Python program (top-code-ans) has a frequency equal to or higher than that of the most frequent answer from the Step-by-Step solution (top-step-by-step-ans), then the top-code-ans is selected as the output. Conversely, top-step-by-step-ans is chosen as the final output.

\section{Experiments}
\begin{table*}[!t]
    \centering
    \setlength\tabcolsep{1.0pt}
    \begin{tabular}{c|cccccccccc}
        \Xhline{1pt}
        \multirow{2}{*}{Method} & \multicolumn{9}{c}{Dataset}  \\
        \cmidrule{2-10}
         & AddSub & SingleEQ & MultiArith & SVAMP & GSM8K & AQuA & Math23K & MathMC & MathToF \\
        \Xcline{1-1}{0.4pt}
        \Xhline{0.4pt}

        Previous SoTA & 95.7 & 98.8 & 100.0 & 93.7 & 97.0 & 79.9 & 87.1 & - & - \\

        \hline
 
        CoT (GPT-3.5-Turbo)  & 89.6 & 90.2 & 95.0 & 82.2 & 75.5 & 60.6 & 63.5 & 60.0 & 68.3\\
        Ours (GPT-3.5-Turbo) & \textbf{92.7} & \textbf{98.2} & \textbf{98.0} & \textbf{86.0} & \textbf{84.3} & \textbf{70.8} & \textbf{88.3} & \textbf{78.8} & \textbf{78.8}\\
        \quad & (+3.1) & (+8.0) & (+3.0) & (+3.8) & (+8.8) & (+10.2) & (+24.8) & (+18.8) & (+10.5)\\

        \hline

        CoT (GPT-4) & 95.7 & 94.5 & 98.6 & 92.6 & 91.2 & 76.4 & 83.2 & 88.1 & 82.5\\
        Ours (GPT-4) & \textbf{98.2} & \textbf{98.6} & \textbf{99.0} & \textbf{93.9} & \textbf{94.8} & \textbf{81.1} & \textbf{94.3}  & \textbf{92.2} & \textbf{89.2} \\
        \quad & (+2.5) & (+4.1) & (+0.4) & (+1.3) & (+3.6) & (+4.7) & (+11.1) & (+4.1) & (+6.7)\\
  
        \Xhline{1pt}
    \end{tabular} 
    \caption{Evaluation results of teaching-inspired integrated prompting framework applied on different models, GPT-3.5-Turbo and GPT-4. Our approach improves performance on different models and benchmarks. Our approach achieves state-of-the-art performance on AddSub, SVAMP, Math23K and AQuA benchmarks on GPT-4. Previous state-of-the-art performance are from \cite{gao2023pal} for SingleEQ, \cite{wang2022self} for MultiArith, \cite{zhao2023automatic} for AddSub and SVAMP, \cite{zhou2023solving} for GSM8K, \cite{zheng2023progressive} for AQuA, \cite{zhang2022multi} for Math23K dataset.}
    \label{tbl:tip_7_datasets}
\end{table*}
\subsection{Experimental Settings}
\noindent\textbf{Datasets.} Our method is evaluated on six English mathematical reasoning benchmarks: AddSub \citep{hosseini2014learning}, SingleEQ \cite{koncel2015parsing}, SVAMP \cite{patel2021nlp}, MultiArith \cite{roy2015solving}, GSM8K \cite{cobbe2021training}, AQuA \cite{ling2017program}, one Chinese math reasoning benchmark, Math23K \cite{wang2017deep}, and two datasets (MathMC and MathToF) we construct.

\noindent\textbf{Models.} For our experiments, we use two LLMs from the GPT series: GPT-3.5-Turbo \cite{ouyang2022training} and GPT-4 \citep{openaigpt}. We conduct all our experiments by employing OpenAI's API, ensuring that our methodology aligns with standard practices and is easy to replicate. 

\noindent\textbf{Prompts and Hyperparameters.} Specific prompts are detailed in Appendix \ref{prompts-appendix}. The prompts include only the most relevant similar problem. For the greedy decoding strategy, the temperature is set to 0.0, while for the Self-Consistency strategy, it is adjusted to 0.5.

\begin{table*}[!t]
    \setlength\tabcolsep{1pt}
    \centering
    \begin{tabular}{c|cccccccccc}
        \Xhline{1pt}
        \multirow{2}{*}{Method} & \multicolumn{9}{c}{Dataset}  \\
        \cmidrule{2-10}
         & AddSub & SingleEQ & MultiArith & SVAMP & GSM8K & AQuA & Math23K & MathMC & MathToF &\\
        \Xcline{1-1}{0.3pt}
        \Xhline{1pt}
        Ours & \textbf{92.7} & \textbf{98.2} & \textbf{98.0} & \textbf{86.0} & \textbf{84.3} & \textbf{70.8} & \textbf{88.3} & \textbf{78.8} & \textbf{78.8} \\
        w/o BG + Sim\_Prob  & 90.3 & 95.4 & 97.2 & 84.7 & 83.4 & 68.5 & 79.6 & 64.4 & 73.0\\
        \quad & (-2.4) & (-2.8) & (-0.8) & (-1.3) & (-0.9) & (-2.3) & (-8.7) & (-14.4) & (-5.8)\\
        w/o Python Codes & 93.4 & 98.2 & 97.8 & 85.7 & 75.2 & - & 84.2 & - & - & \\
        \quad & (+0.7) & (+0.0) & (-0.1) & (-0.3) & (-9.1) & - & (-4.1) & - & -\\
        w/o Ans\_Selection & 91.4 & 91.3 & 95.0 & 83.6 & 75.4 & 60.6 & 76.5 & 70.6 & 74.3 & \\
        \quad & (-1.3) & (-6.9) & (-3.0) & (-2.4) & (-8.9) & (-10.2) & (-11.8) & (-8.2) & (-4.5)\\
        w/o Eng\_Translate & - & - & - & - & - & - & 88.5 & 75.8 & 77.8 \\
        \quad & - & - & - & - & - & - & (+0.2) & (-3.0) & (-1.0)\\
        \Xhline{1pt}
    \end{tabular}
    \caption{Ablation study for different components of our proposed integrated prompting framework. The study utilized seven public datasets and two datasets we created, with all experiments on the GPT-3.5-Turbo model.}
    \label{tbl:ablation_study_integrated}
\end{table*}

\subsection{Main Results}

We compare the evaluation results of our integrated prompting framework with the Chain-of-Thought method on GPT-3.5-Turbo and GPT-4 models. As shown in Table \ref{tbl:tip_7_datasets}, our framework improves the mathematical reasoning performance significantly over two models on seven math datasets, especially improving 8.8\% on GSM8K, 24.8\% on Math23K, 8.0\% on SingleEQ and 10.2\% on AQuA when used on GPT-3.5-Turbo. Surprisingly, with GPT-4 and our integrated prompting framework, we achieve the new state-of-the-art performance on four math benchmarks (AddSub, SVAMP, Math23K and AQuA) with accuracies of 98.2\% (+3.3\%), 93.9\% (+0.2\%), 94.3\% (+7.2\%) and 81.1\% (+1.2\%). 

Additionally, we present the results of GPT-3.5-Turbo and GPT-4 on two datasets we created, MathMC and MathToF. As seen in Table \ref{tbl:tip_7_datasets}, leveraging our prompt method on GPT-3.5-Turbo yields a significant enhancement in performance, boosting reasoning accuracy on MathMC by 18.8\% and MathToF by 10.5\%, respectively. On deploying the same prompting framework on GPT-4, we observed a marked improvement as well, with increases of 4.1\% and 6.7\% in respective metrics. These results demonstrate the efficacy of our approach in facilitating reasoning tasks.
\subsection{Ablation Study}

\subsubsection{Similar problems and background knowledge.} Removing similar problems and background knowledge from the prompts leads to a general decline in accuracy across nine datasets shown in Table \ref{tbl:ablation_study_integrated}. This indicates that similar problems and background play a guiding role in enhancing LLM reasoning.

\subsubsection{Python programs and double-check verification strategy.}
The results in Table \ref{tbl:ablation_study_integrated} demonstrate that removing the Python program generation and double-check strategies had minimal impact or even a slight improvement in accuracy for simpler math problem sets (AddSub, SingleEQ and MultiArith). However, for more complex problem sets (GSM8K, Math23K), there is a noticeable decrease in accuracy by 9.1\% and 4.1\% respectively. This indicates that incorporating Python program-generated answers and the double-check strategy helps compensate for LLMs susceptibility to computational errors in complex calculations.
\subsubsection{Answer Selection strategy.}Analyzing the experimental results, it can be found that the accuracy decrease among nine datasets, especially on more complex datasets including GSM8K, AQuA, and Math23K, where the accuracy dropped by 8.9\%, 10.2\%, and 11.8\% respectively. When combined with self-consistency and double-check-verification methods, simple calculation errors or occasional faulty reasoning can be avoided. Different problem-solving paths and calculation methods (Python program or natural language) would produce the same results for a given problem.
\subsubsection{English-Chinese-Ensembling strategy.} We evaluate the impact of the English-Chinese Ensembling strategy on three Chinese datasets. When we removed this component, the accuracy on MathMC and MathToF dropped by 3.0\% and 1.0\%. This finding suggests that translating Chinese problems into English can make it easier for the language model to understand, thereby generating more accurate solutions.
\subsubsection{The number of similar problems.}
\begin{figure}[!t]
\centering
\includegraphics[width=\columnwidth]{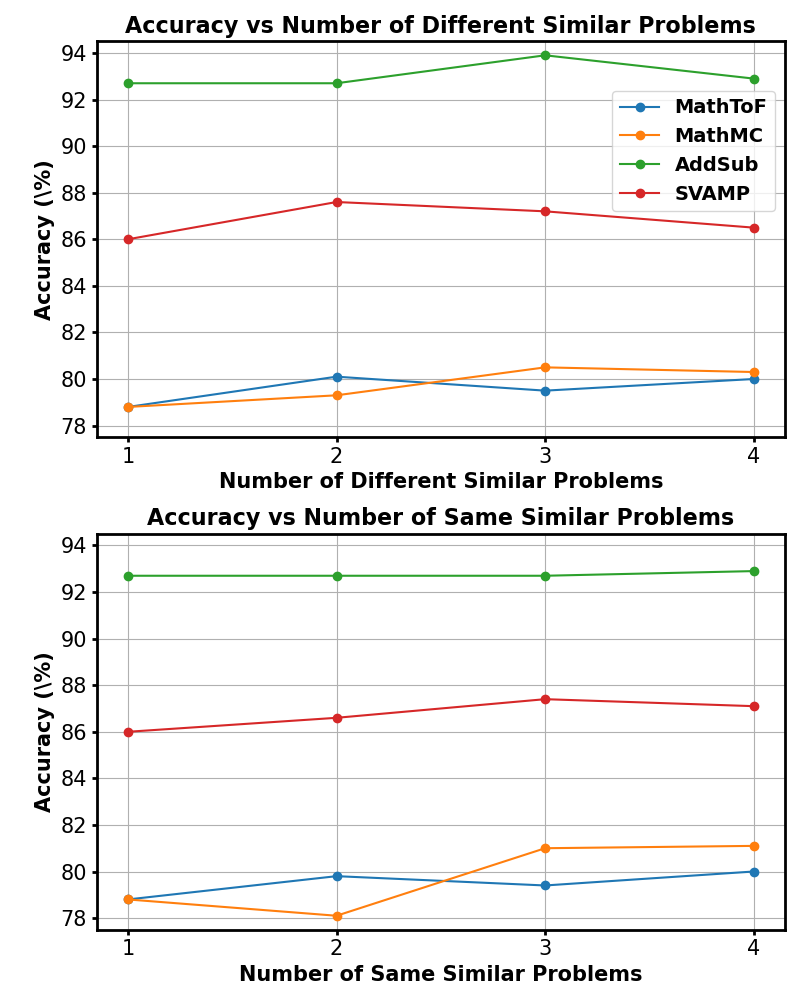} 
\caption{Experiment results of adding different number of varying or the same similar problems.}
\label{fig-ab-pro}
\end{figure}

We explore the affect of the number of similar problems by adding different number of varying or the same similar problems\footnote{Adding K number of the same similar problems refers to repeating the exact  exact same similar problem K times within the prompt.} to prompts. Figure \ref{fig-ab-pro}(a) shows that the effectiveness of adding similar questions is not solely determined by quantity. When the added questions differ significantly from the target question, it can negatively affect accuracy. However, within a certain similarity threshold, increasing the quantity of similar questions improves LLMs reasoning accuracy. Figure \ref{fig-ab-pro}(b) demonstrates that including the multiple same top-similar questions in prompts leads to an notable improvement in LLMs reasoning accuracy. This approach indirectly addresses the challenge of acquiring external information, benefiting LLMs in capturing and utilizing relevant external information more effectively. 
\section{Conclusion}
This paper presents an innovative teaching-inspired integrated prompting framework, to conquer the limitations of LLMs in arithmetic reasoning tasks. The framework emulates the teaching process to introduce essential concepts, theorems, and similar problems to LLMs, and designing double-checking and answer selection mechanisms thereby significantly enhancing their capability to perform arithmetic reasoning tasks. Empirical results reveal that employing our framework leads to substantial improvements in arithmetic reasoning accuracy. Our study also underscores the need for more diverse and comprehensive benchmarks for evaluating the performance on arithmetic reasoning, which we address by introducing the MathMC and MathToF datasets. 
In future work, researchers can further refine and explore its applicability to other domains.


\bibliography{custom}

\appendix
\appendix
\section*{Appendix}
\section{Similar Problems and Background Knowledge Retrieval}

\begin{figure*}[!t]
\centering
\includegraphics[trim=5cm 10cm 5cm 5cm,width=0.5\textwidth]{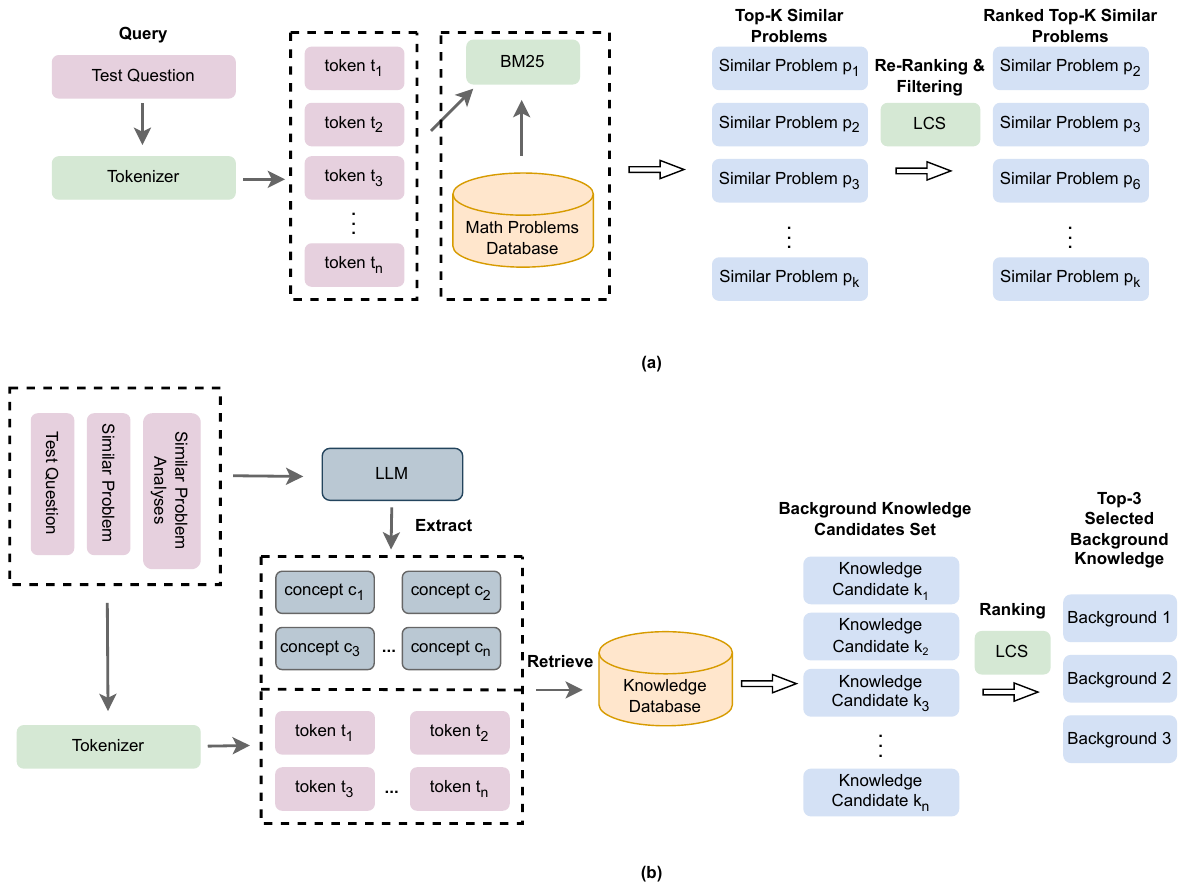} 
\caption{The procedure of similar problems retrieval (a) and the background knowledge generation (b).}
\label{fig_sim_bg}
\end{figure*}

Figure \ref{fig_sim_bg} illustrates the process of similar problems and background knowledge retrieval. 

For background knowledge database, we curated a rich repository from mathematical textbooks, exercise workbooks, and web sources. From these resources, we extract a wealth of knowledge points and background information, encompassing theories, theorems, and problem-solving methodologies. Subsequently, we compile and store these findings to establish a comprehensive knowledge base.

In parallel, the similar problems database comprises problems sourced from mathematical textbooks and workbooks, each accompanied by detailed analyses derived from these materials.

\section{Dataset Details and Sample Questions}\label{data-appendix}
\begin{figure*}[!t]
\centering
\includegraphics[trim=8cm 0cm 8cm 0cm,width=1\columnwidth]{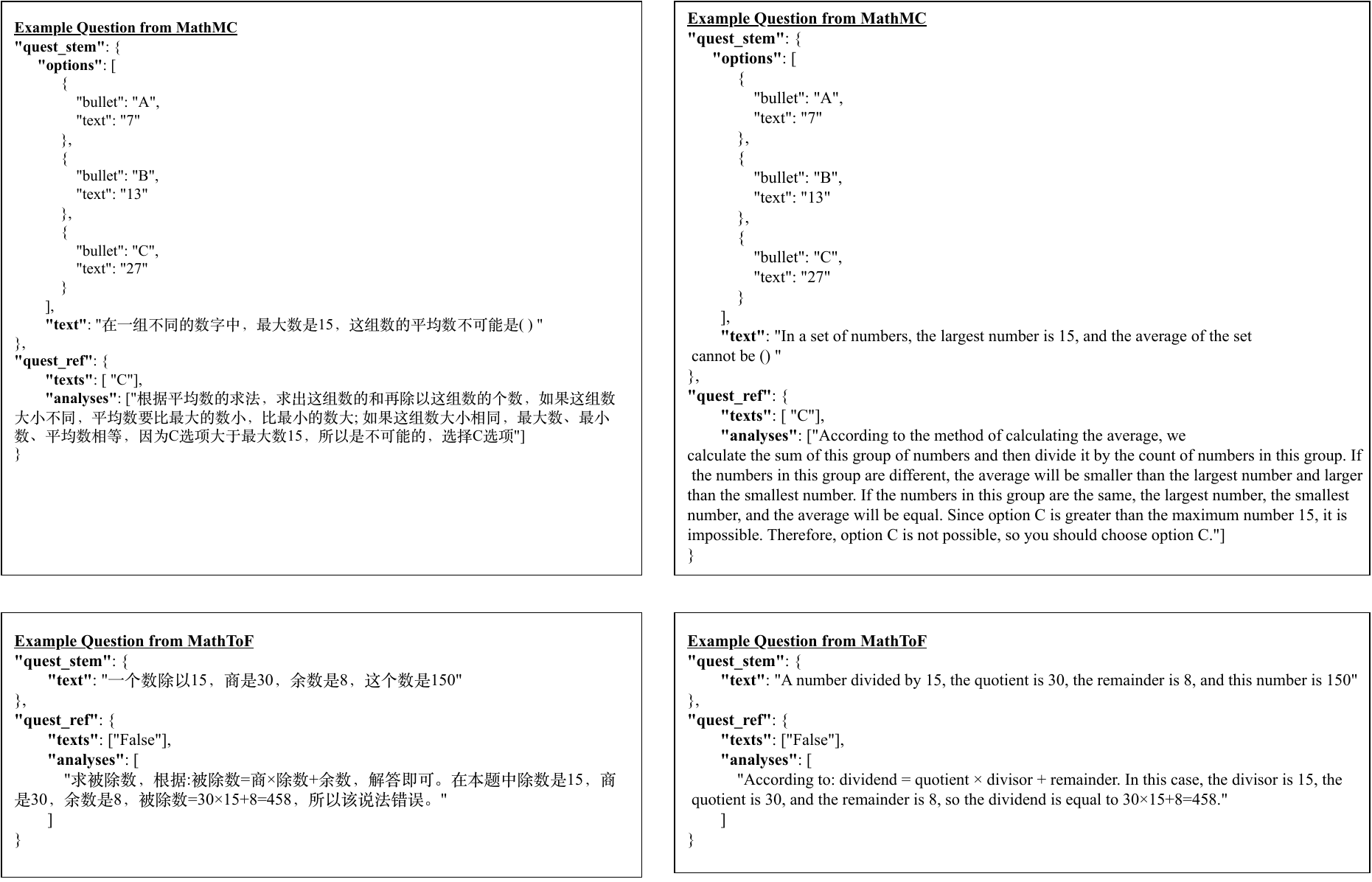} 
\caption{Sample questions of MathMC and MathToF datasets. The left side displays the original questions in Chinese, while the right side shows the same questions translated into English}
\label{fig_appendix}
\end{figure*}
In this section, we discuss the details and specific segments of the two datasets we have established. Figure 5 showcases two sample questions from the dataset, one being a multiple-choice question, and the other a true-or-false question.

Each multiple-choice question is divided into two segments. The first segment, the question stem, is composed of the question itself and the possible choices. The second segment includes the correct answer and a comprehensive explanation of the solution. Similarly, the structure of the true-or-false questions includes a question stem, the correct answer (either True or False), and a detailed rationale.
\section{Prompts}\label{prompts-appendix}
In this section, the System Prompt, Teaching-Inspired Prompt, Standard Prompt and Chain-of-thought Prompt we used in our experiments are shown in the following. 
\subsection{System Prompt}
You are an super smart elementary school math teacher. You need read the math problem carefully and solve it in a step by step way to be sure you have the right answer. You often make mistakes in calculations, so please be careful when calculating.

Please do not be influenced by the typos in the question and reason based on the semantics of the question.

Please make sure your replies as simple and easy to understand as possible.
\subsection{Teaching-Inspired Prompt}
This section shows the Teaching-Inspired Prompting format and one example of our proposed prompt. 
\noindent\\
If there is a reference question and the reference question is very similar to the question you need to answer, you should think based on the analysis process of the reference question, but you cannot be affected by its question stem.You still need to return the complete analysis process of the question you need to answer. \\
Reference question: sim\_stem \\
Reference analysis: sim\_analysis \\
Reference answer: sim\_ans \\

\noindent You may use the following background knowledge when analyzing the problem: \\
Background: background \\
Question: question to be solved\\
\subsubsection{Example}
\noindent\\
If there is a reference question and the reference question is very similar to the question you need to answer, you should think based on the analysis process of the reference question, but you cannot be affected by its question stem.You still need to return the complete analysis process of the question you need to answer. \\
Reference question: In Class 6, there are a total of 52 students. Among them, 30 students like to eat rice, and 29 students prefer noodles. The number of students who like both rice and noodles is ( ). \\
Reference analysis: Based on the information "There are a total of 30 students who like to eat rice and 29 students who prefer noodles," we can calculate the total number of students who like either rice or noodles: 30 + 29 = 59. However, this count includes the students who like both rice and noodles twice. Therefore, applying the principle of inclusion-exclusion, we can determine that the number of students who like both rice and noodles is 59 - 52 = 7. Thus, the answer is 7.
Reference answer: 7 \\
\noindent You may use the following background knowledge when analyzing the problem: \\
Background: principle of inclusion-exclusion: $|A \cup B|=|A|+|B|-|A \cap B|$ \\
Question: In order to prepare for the fruit party, Class 3 made statistics on the two kinds of fruits that everyone liked. 38 students like to eat bananas, 32 students like to eat fragrant pears, and 10 students like both. How many students in Class 3?

\subsection{Chain-of-Thought Prompt}
\subsubsection{Word Problems}
\noindent\\
\textbf{a) Few-Shot Examples}\\
Examples: \\
Question: Xiaoming is 5 years old this year, and his father is 25 years old this year. How old will Xiaoming be when his father is 30 years old? \\
Analysis: \\
thought: \\
When the father is 30 years old, 5 years have passed since he was 25. \\
At this time, Little Ming should be 10 years old (5 + 5). \\
steps: \\
1. We need to figure out how many years it will take for the father to reach 30 years old from now (25 years old). This can be obtained by subtracting 25 from 30, that is, 30-25=5.  Therefore, the father still needs 5 years to reach 30 years old. \\
2. We know that Little Ming is now 5 years old, so his age will increase in the next 5 years. Since his age increases by 1 year every year, in 5 years his age will increase by 5 years. \\
3. If we add Little Ming's current age of 5 to the increase of 5 years in the next 5 years, we can get Little Ming's age when his father is 30 years old. That is, 5+5=10. \\
answer: \\
10 \\

\noindent Question: Xiaoming read 30 pages on the second day, and
read one more page than the second day on the first day.
How many pages did he read on the first day? \\
Analysis: \\
thought: \\
Since Xiaoming read 30 pages on the second day and read one more page than the second day on the first day, Xiaoming read 31 pages on the first day. \\
steps: \\
1. Xiaoming read one more page on the first day than on the second day. \\
2. Xiaoming read 30 pages on the second day. \\
3. Therefore, the number of pages Xiaoming read on the first day is one more than that of the second day. \\
4. Thus, Xiaoming read 30 pages + 1 page on the first day, which is equal to 31 pages. \\
answer: \\
31 pages \\
\\
\textbf{b) Reply Format}\\
When you are certain that the answer is correct, you need to return the following content: \\
thought: [Return your thinking process for solving this problem.] \\
steps: [Return the detailed solution steps.] \\
answer: [The answer to the question. If there are multiple questions in the problem, the answer format should be: (1) Answer to the first question. (2) Answer to the second question....] \\
Important: Your return format must be consistent with the Examples \\
Important: The content you return must include fore keywords: thought, steps, and answer, and the content of every keyword cannot be empty. Besides, every keyword should be in English. \\

\subsubsection{Multiple-Choice Problems}
\noindent \\
Examples: \\
Stem: The approximate distance from Xiao Ning's home to school, given that he walks an average step length of 58 centimeters and has taken 135 steps, is about () \\
Options: A.8000m B.80m C.70m \\
thought: Based on the formula distance = number of steps × length per step, write the equation 58 × 135, calculate it using integer multiplication method, and get the result of 7830. Then, according to the rounding rule, the answer can be solved. \\
steps: \\
1. Using the formula distance = number of steps × length per step, derive the equation 58 × 135 \\
2. According to the equation 58 × 135 = 7830 cm, determine the distance from Xiao Ning house to the school as 7830 cm \\
3. Since the options are in meters and the result we calculated earlier is in centimeters, we should convert centimeters to meters. 7830 cm = 78.3 m \\
4. Applying rounding rules, 78.3 m is approximately equal to 80 m, so option B should be selected \\
answer: B

\noindent Stem: Which of the following statements is correct?  \\
Options: A. A ray is 50 meters long B. There are 6 big months (31 days) and 6 small months (30 days) in a year C. 1/3:1/4 and 4:3 can form a proportion D. The whole year in 2020 has 365 days. \\
thought: Detemine four choices of ABCD in the question are correct or not \\
steps: \\
1. Option A, since a ray has only one endpoint and extends infinitely in one direction, it cannot be measured in terms of length. Therefore, Option A is incorrect. \\
2. Option B, there are 7 big months and 5 small months in a year, so the statement in Option B is incorrect. \\
3. Option C, to form a proportion, the ratios on both sides should be equal. 1/3:1/4 = 4:3 = 4/3, and 4:3 is equal to 4/3. Therefore, it can form a proportion with 4:3. The statement in Option C is correct. \\
4. Option D, 2020 is a leap year because it is divisible by 4, so the whole year has 366 days. The statement in Option D is incorrect. \\
5. Therefore, the correct answer is Option C. \\
answer: C

\noindent Stem: Which of the following expressions has a value greater than 100?  \\
Options: A.50+45 B.90+20 C.90-80 \\
thought: Compare the result of adding each equation to 100. \\
steps: \\
1. The result of option A is 50 + 45 = 95, which is less than 100, so Option A is incorrect. \\
2. The result of option B is 90 + 20 = 110, which is greater than 100, so Option B is correct. The correct answer is B. \\
3. To prevent calculation errors, let us calculate the answer for Option C again. 90 - 80 = 10, which is less than 100, so Option C is also incorrect. \\
4. Therefore, the final answer is B. \\
answer: B\\
\\
\textbf{b) Reply Format}\\
When you are certain that the answer is correct, you need to return the following content:\\
thought: It's necessary. Return your thinking process for solving this problem. \\
steps: It's necessary. The steps for solving the question, with as much detail as possible. \\
answer: It's necessary. The specific option to the question, such as A/B/C/D \\
Important: Your return format must be consistent with the Examples \\
Important: The content you return must include the keyword: thought, steps and answer and the content of every keyword cannot be empty. Besides, each keyword should be in English. \\

\subsubsection{True-or-False Problems}
\noindent \\
Examples:
Question: True or False: The number that is 100 more than the largest three-digit number is 1999. \\
'thought: Firstly, we need to know what the largest three-digit number is, and then calculate the largest three-digit number plus 100 to determine whether the result is equal to 1999. If the result is not equal to 1999, then the statement is false. If it is equal to 1999, then the statement is true. \\
steps: \\
1. The largest three-digit number is 999. \\
2. Adding 100 to 999 results in 1099. \\
3. The result of the calculation is 1099, which is not equal to 1999. Therefore, the answer to this question is false. \\
answer: False \\
\\
Question: True or False: The '9' in 0.019 is in the hundredths place. \\
thought: The first decimal place to the right of the decimal point is the tenths place, the second decimal place is the hundredths place, and the third decimal place is the thousandths place. \\
steps:  \\
1. To determine the hundredths place, we need to look at the second decimal place to the right of the decimal point. \\
2. Looking at the third decimal place to the right of the decimal point in 0.019, we find that it is 9. \\
3. We can conclude that the "9" in 0.019 is in the thousandths place. \\
4. Therefore, the statement in the question is false. \\
answer: False \\
\\
Question: True or False: The remainder is never greater than the quotient.\\
thought: This statement can be judged by the relationship between the remainder, divisor, and quotient, or by giving examples to see if the statement is true or false.\\
steps: \\
1. Generally, the remainder cannot be greater than the divisor, but there is no absolute relationship between the remainder and the quotient.\\
2. For example, 104 divided by 33 equals 3 with a remainder of 5, where the remainder 5 is greater than the quotient 3.\\
3. Another example is 3 divided by 4, which equals 0 with a remainder of 3, where the remainder 3 is greater than the quotient 0.\\
4. Therefore, based on the counterexamples and concept relationships, we can conclude that this statement is false.\\
5. Therefore, the final answer is false.\\
answer: False\\
\\
\textbf{b) Replay Format}\\
When you are certain that the answer is correct, you need to return the following content: \\
thought: It's necessary. Return your thinking process for solving this problem.\\
steps: It's necessary. The steps for solving the question, with as much detail as possible. \\
answer: It's necessary. If you believe that the statement in the question is correct, return True. If you believe that the statement in the question is false, return False. \\
Important: Your return format must be consistent with the Examples \\
Important: The content you return must include the keyword: thought, steps and answer. and the content of every keyword cannot be empty. Besides, each keyword should be in English. \\

\end{document}